\documentclass[preprint,12pt]{elsarticle}
\pdfoutput=1

\makeatletter
\def\ps@pprintTitle{%
 \let\@oddhead\@empty
 \let\@evenhead\@empty
 \def\@oddfoot{\centerline{\thepage}}%
 \let\@evenfoot\@oddfoot}
\makeatother

\usepackage{gensymb}
\usepackage{amssymb}
\usepackage{amsmath}
\usepackage{caption}
\usepackage{subcaption}
\usepackage{lineno}
\usepackage{tabularx}

\usepackage{booktabs}
\usepackage{multirow}
\usepackage[margin=2.5cm]{geometry}
\usepackage{array}
\newcolumntype{C}[1]{>{\centering\arraybackslash}p{#1}}
\newcolumntype{L}{>{\centering\arraybackslash}m{4cm}}
\newcolumntype{M}{>{\centering\arraybackslash}m{3cm}}
\newcolumntype{N}{>{\centering\arraybackslash}m{2cm}}

\usepackage[ruled,vlined]{algorithm2e}

\begin{document}

\begin{frontmatter}

\author[inst1]{Arsenii Uglov\corref{cor1}}
\corref{cor1}
\ead{urrusmsng@gmail.com}

\author[inst1]{Sergei Nikolaev\corref{cor2}}
\corref{cor2}
\ead{s.nikolaev@skoltech.ru}

\author[inst1]{Sergei Belov}
\author[inst1]{Daniil Padalitsa}
\author[inst1]{Tatiana Greenkina}

\author{Marco San Biagio}\corref{cor3}
\corref{cor3}
\ead{marco.sanbiagio@gmail.com}
\author[inst2]{Fabio Cacciatori}\corref{cor4}
\corref{cor4}
\ead{cacciatori@illogic.xyz}

\affiliation[inst1]{organization={Skolkovo Institute of Science and Technology},
            addressline={Bolshoy Boulevard 30, bld. 1},
            city={Moscow},
            postcode={121205},
            country={Russia}}

\affiliation[inst2]{organization={Illogic s.r.l},
            addressline={Corso stati Uniti 57,10128},
            city={Turin},
            country={Italy}}

\title{Surrogate Modelling for Injection Molding Processes using Machine Learning}

\begin{abstract}
Injection molding is one of the most popular manufacturing methods for the modeling of complex plastic objects. Faster numerical simulation of the technological process would allow for faster and cheaper design cycles of new products. In this work, we propose a baseline for a data processing pipeline that includes the extraction of data from Moldflow simulation projects and the prediction of the fill time and deflection distributions over 3-dimensional surfaces using machine learning models. We propose algorithms for engineering of features, including information of injector gates parameters that will mostly affect the time for plastic to reach the particular point of the form for fill time prediction, and geometrical features for deflection prediction. We propose and evaluate baseline machine learning models for fill time and deflection distribution prediction and provide baseline values of MSE and RMSE metrics. Finally, we measure the execution time of our solution and show that it significantly exceeds the time of simulation with Moldflow software: approximately 17 times and 14 times faster for mean and median total times respectively, comparing the times of all analysis stages for deflection prediction. Our solution has been implemented in a prototype web application that was approved by the management board of Fiat Chrysler Automobiles and Illogic SRL.
As one of the promising applications of this surrogate modelling approach, we envision the use of trained models as a fast objective function in the task of optimization of technological parameters of the injection molding process (meaning optimal placement of gates), which could significantly aid engineers in this task, or even automate it.
\end{abstract}



\end{frontmatter}

\section{Introduction}

Injection molding is a widespread manufacturing process for producing objects of given shapes from various materials. It is the most common modern method of manufacturing of plastic parts, and it is ideal for producing high volumes of copies of the same object. Numerical analysis is crucial for designing products with given mechanical properties and a minimal number of defects. The industry standard is to use extensive fluid-dynamics-based simulations with specialized software such as Autodesk Moldflow. This kind of simulation is computationally quite burdensome and time-consuming. As well, the final result is highly dependent on the experience and skills of the engineers, introducing variance in quality. Some manufacturers are interested in shorter and cheaper design cycles with less human interaction.

One way to accelerate a design cycle and expand searchable design space is to implement surrogate-based optimization  \cite{han_surrogate-based_2012}. Data-driven models such as Gaussian processes, support vector machines (SVM) and artificial neural networks (ANN) are common choices for surrogate models.

In recent years, we have observed a steady rise in interest in the application of deep learning (DL) models to model physical or engineering processes. For example, generative adversarial networks (GANs) are used in high energy-physics for fast approximate simulation of detector responses \cite{ratnikov_generative_2020, ratnikov_using_2020}, and supervised DL architectures are used in geology for real-time simulation of seismic waves \cite{moseley_fast_2018} and for semiconductor manufacturing  \cite{orihara_approximation_2018}.

In this work we consider a deep learning approach to creating a surrogate model, based on Moldflow simulation data, which predicts the distribution of fill time and deflection values over the 3d surface as targets of the simulation.

Our research into related works on this topic has, as of 2020, found no description of a similar approach for the described task, so we consider our results to be a baseline.

This work was created as a part of a project with an Italian automotive industry partner Fiat Chrysler Automobiles. They provided us with a dataset of Moldflow project files of simulations. Since this work is covered by the non-disclosure agreement, including the dataset, we cannot disclose the partner’s name or provide many vivid data representations, as the format of the data suggests. However, we will provide summary details of a dataset, a description of the used features, details of our data processing pipeline and machine learning (ML) models, and the achieved values of the result metrics.

\section{Injection Molding Process}

\subsection{Molding Process}

\begin{figure*}[h]
    \centering
    \includegraphics[width=0.7 \textwidth]{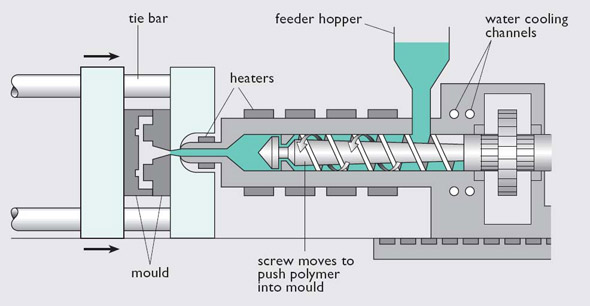}
    \caption{Injection molding machine, picture takes from \cite{mould_machine}}
    \label{fig:machine}
\end{figure*}

An injection molding machine (Figure \ref{fig:machine}) consists primarily of two units: a clamping unit and an injection unit. The clamping unit opens and closes the die and maintains the holding pressure. The injection unit melts plastic pellets and creates melted plastic flow under either controlled speed or pressure (depending on the stage of the process). It consists of a hopper, heaters and a screw which propels the plastic flow. Speed and pressure are controlled by the rotation speed of the screw.

The mold is usually a steel cube split into two halves with cavities in the shape of the product and auxiliary cavities for plastic canals and a cooling system.

A typical process of injection molding consists of the following steps:

\begin{enumerate}
    \item Clamping: closing the halves of the mold
    \item Injection: introduction of melted polymer flow into the cavity
    \item Dwelling: equalizing of polymer pressure through the whole cavity
    \item Cooling
    \item Mold opening
    \item Product ejection
\end{enumerate}

The product’s shape is produced by an industrial designer in the form of a CAD file. The engineer considers the given shape and proposes technical parameters for the molding machine and corrections to the shape if needed. To do so, the engineer iteratively selects parameters based on his or her domain knowledge and tests it with numerical analysis of the whole molding process in specialized software such as Epicor, Moldex3D or Autodesk Moldflow.

In this work we consider complex thin-walled plastic parts such as car dashboards and bumpers.

\subsection{Numerical Simulation}
Simulation allows us to predict such things as fill time, cooling time, and number and type of defects based on technical parameters such as material, mold temperature, cooling system, flow speed, pressure profiles, and geometry itself.

For thin-walled shapes Moldflow offers the so-called dual domain method of modeling. The analysis takes place on the surface of the mold cavity expressed as triangular mesh. Special connector elements are used to synchronize the results on the opposing faces.

A typical simulation sequence is Fill + Pack + Warp + Cool analysis. Fill analysis models polymer flow during the injection stage. Pack analysis models the same during the dwelling stage. Cool analysis models heat exchange between elements of the mesh, the mold and the cooling system. Warp analysis predicts the types and locations of defects. One of the most important results of analysis is spatial distribution of deflection, or displacement from the set shape under the load, in the final product. During simulations MoldFlow solves a system of differential equations that describes the fluid dynamics and heat exchange on the surface of the mesh using the finite elements method.

\begin{figure*}[h]
    \centering
    \includegraphics[width=0.8 \textwidth]{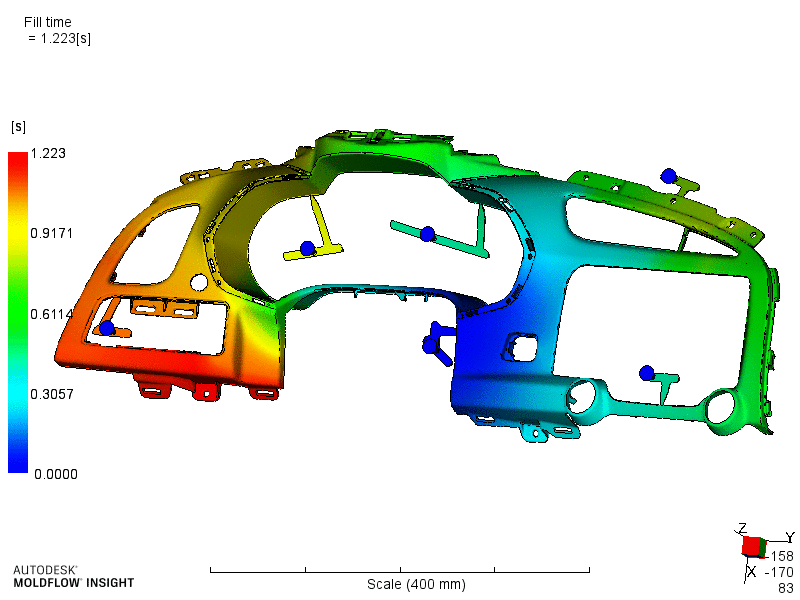}
    \caption[test]{Moldflow prediction of fill time for generic car dashboard. Image is taken from an open information source \cite{Muhendislik}}
    \label{fig:filltimel}
\end{figure*}

As input parameters for a simulation scenario we can, among others, emphasize these technological parameters: mold geometry, type of plastic, position of input gates and cooling channels, temperature of plastic and mold, opening time of each gate, flow pressure, and clamping pressure. The results of simulation are spatial distribution of fill time for Fill analysis, cooling time for Cool analysis, and weld lines and deflection for Warp analysis. Figure \ref{fig:filltimel} shows an example of output of the fill time simulation for a generic dashboard model.

\section{Related Works}
Here we review the most common approaches that apply machine learning models to three-dimensional objects. Most models can be distinctively categorized by used representation of 3D objects.

\subsection{Meshed Representation}
Meshes discretize an object's surface with a set of flat polygons, usually triangles. It is the most common representation in computer graphics and CADs.

The simplest approach to deal with the mesh would be to treat it as a graph and encode geometrical properties into vertex and edge signals. The two main approaches to constructing graph convolutions are based on spectral filtering or local filtering. Depending on the type of specter approximation or type of local filters, different operations can be derived. Nonetheless, most models allow general representation in terms of a message passing to a node from its neighborhood.

Graph networks are usually isotropic in terms of mutual orientation of nodes and are invariant under permutations of the neighborhood, but mutual orientation of nodes in a mesh is crucial to defining that mesh. Convolutional operators should thus be anisotropic in terms of orientation between the target and source nodes. However, any definition of orientation via angle would arbitrarily depend on the choice of reference angle. We thus either need to fix canonical orientation in the node neighborhood for all meshes, as it is done in the case of image convolutions on a regular grid. In \cite{gong_spiralnet_2019}, the authors consider similarly meshed surfaces and introduce fixed serialization of the neighborhood by spiral trajectory from the target; the authors of \cite{verma_feastnet_2018} use an attention mechanism for soft assignment of neighbors to kernel weight maps.

As stated by Bernstein et al. in \cite{bronstein_geometric_2017}, the success of convolution on regular data is based on implicitly exploited global symmetries of Euclidean spaces. 2-d convolutions, for example, are naturally shift equivariant. A general manifold, and meshes in particular, lacks global symmetries that can be exploited, but one can define some locally equivariant operations.
Cohen et al. in \cite{de_haan_gauge_2020} and \cite{cohen_gauge_2019} introduce a specific type of features on node tangent space and derive gauge equivariant convolution.

\subsection{Point Cloud Representation}
In this approach, a 3D object expressed as a set of points in a 3D space sampled from the volume of the object or from the nodes of 3d mesh polygons. It is usually produced by a 3D sensing method such LiDAR scanners or stereo reconstruction. One of the first successful DL models for point clouds was PointNet from  \cite{garcia-garcia_pointnet_2016}, where the same (weights sharing) network is applied to features of each point, and then aggregated by a permutation invariant operator to produce embedding of a whole cloud. Such an approach cannot grasp the local geometry of an object’s surface. In \cite{qi_pointnet_2017} Qi et al. proposed clustering a point cloud and applying PointNet to each cluster, resulting in new point cloud, which is processed in same fashion until the whole point cloud does not merge.

In \cite{wang_dynamic_2019}, the authors dynamically built a k-Nearest Neighbors graph in the feature space of each layer and then applied a graph convolutional network to obtain new features, repeating this process until the desired depth was reached.

Although point cloud models are simpler and can be implemented more easily than meshed models, one of their main drawbacks is that they cannot differentiate between points that are close in 3D space and far in terms of geodesic distance (distance on the surface of the shape).

\subsection{Rasterized representation}
The most common regular approach is to build voxel representation. The object and bounding space around it are divided into a three-dimensional grid of voxels. Such representation allows for trivial extension of 2D convolutions \cite{wu_3d_2015, maturana_voxnet_2015}. However, complexity and memory consumption for such an approach grows cubically with resolution. Effective data structure allows for better time and memory complexity. In \cite{wang_o-cnn_2017}, Wang et al. split bounding space recursively in octets of smaller cubes, and use an oct-tree as an effective storage. Then convolutions act on neighboring cubes on the same level of the tree, passing aggregated values to the coarser level.

Another approach would be in using 2d projections of 3d structures and then applying traditional 2d convolutions. There can be single or multiple views projections of a 3d object with additional 2d features, like the distances to the projection plane, and with some mechanism for aligning predictions from different projections \cite{su2015multi}. Such an approach then allows use of state-of-the-art 2d convolutional architectures for processing of obtained regular 2d representations. And the models used in practice are relatively simple for implementation and training, and they are cost-effective in inference.

It is not quite obvious which of these representations is better suited for our task. In this paper we present the results for 2d projection representation as a baseline.

\section{Our Approach}
\subsection{Data extraction} \label{ch:data_preparation}

The data for further analysis were extracted from Moldflow simulations projects. For this purpose we export simulation data from Moldflow and then, using developed data loader scripts, we extract valuable data from an exported set of files for a particular simulation, preprocess it, and pack the aggregated data for all simulations in the Pandas dataframe. For each simulation we use the following files to extract data from:

\begin{enumerate}
    \item a .pat file that contains mesh data. We collect nodes ids, their coordinates and information for polygons as nodes ids triplets.
    \item a .txt file that contains a Moldflow log file. From this file we extract values of technological parameters (such as Melt temperature, Cooling time, Duration, Filling pressure, Ambient temperature, Mold temperature, Time at the end of filling), and injection gates information (node ids and Opening time).
    \item an .xml file that contains Moldflow simulation results.
\end{enumerate}

Also during the loading process we calculate the shortest paths from nodes to gates considering polygonal mesh as a graph of nodes and edges weighted by the length of the edges. The shortest paths between all nodes and all gates then are calculated according to Dijkstra's algorithm.

\subsection{Data processing pipeline}
\begin{figure*}[h]
    \centering
    \includegraphics[width=0.7\textwidth]{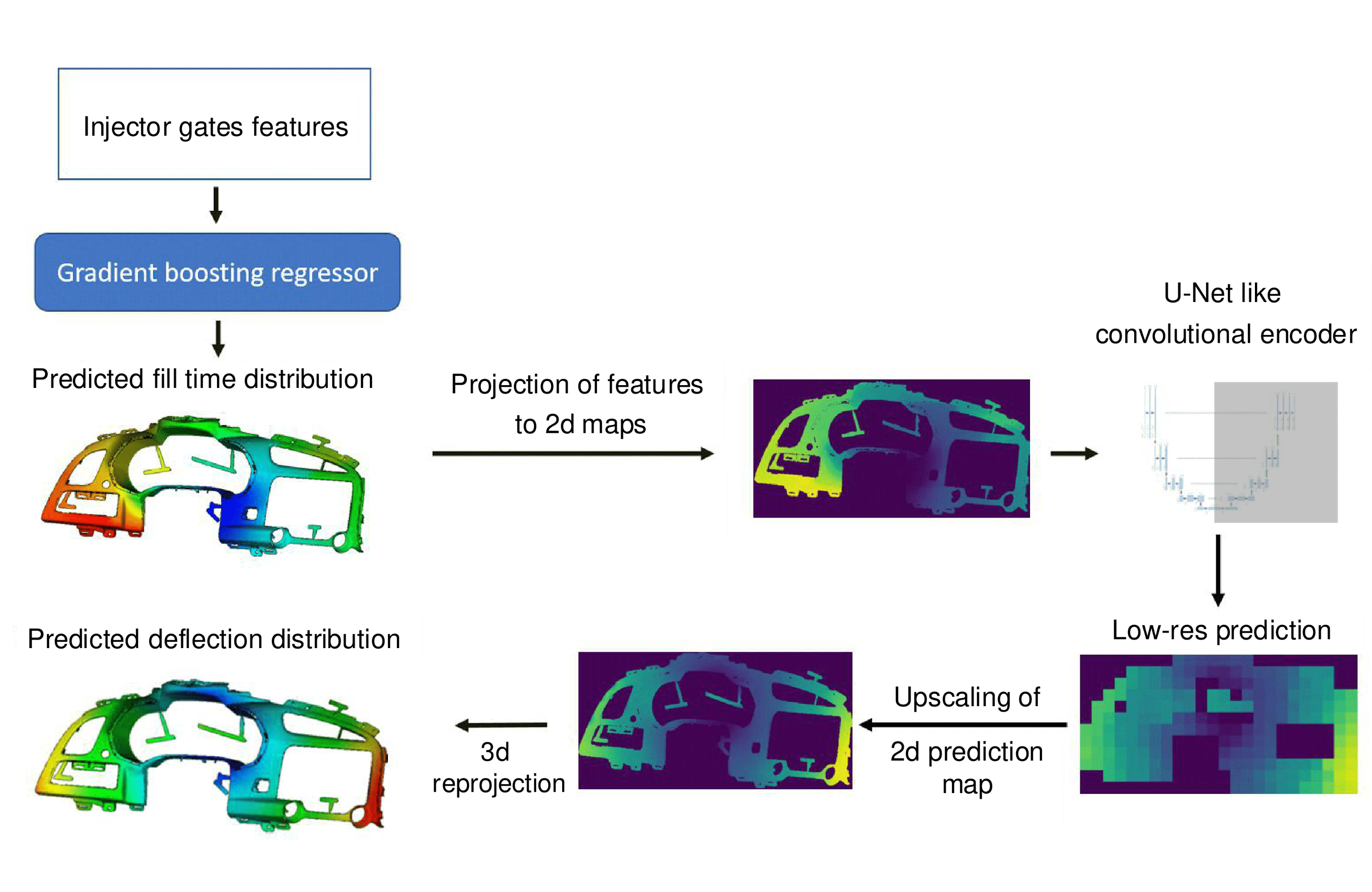}
    \caption{The general scheme of data processing pipeline}
    \label{fig:general_pipeline}
\end{figure*}

The general scheme of our pipeline for the analysis of extracted data is shown on Fig. \ref{fig:general_pipeline}. This diagram shows the general pipeline idea without any actual result distribution or a 3d geometry from our dataset, since we can't show these details due to non-disclosure agreement.

The pipeline has the following sequential steps:
\begin{enumerate}
    \item Pre-process point cloud data
    \item Extract features for gates
    \item Train the Gradient boosting model to predict fill time
    \item Generate 2d feature maps from the 3d model with values
    \item Train a CNN model to predict deflection distribution in 2d space
    \item Reproject 2d deflection distribution back over the 3d mesh
\end{enumerate}

\subsection{Point cloud pre-processing}
At first the subset of points of a given size is randomly selected from the model. We take of about $1/8$ of the total points count.

For each of the selected points we then take k nearest points for the area that will be used for smoothing the prediction results. In our experiments we take k=100.

\subsection{Extraction of features of gates}
For each point we construct an N-dimensional vector of features of gates that will be used to predict fill time value. Since fill time value is dependent on the distance a plastic should travel to reach a particular place, in our experiments for each point we construct features from the three nearest gates, which will mostly affect the time of the plastic to reach this point. That gives us an 8-dimensional feature vector.

As a first set of values, we take distances to the three nearest gates, calculated as described in section \ref{ch:data_preparation}.

As a second triplet of features for each of the three selected gates we use its opening time value.

Finally, we calculate the three-dimensional vectors of orientation of each selected gate relatively to each point and then take the values of $cos(\alpha_1)$, $cos(\alpha_2)$, as the two last features, where $\alpha_1$ and $\alpha_2$ are the angles between the orientation vectors of the first and second and the first and third gates, respectively.

\subsection{Fill time prediction with the Gradient boosting model} \label{ch:filltime_prediction_algo}

Using derived feature vectors for points, we trained the Gradient boosting model to predict the fill time value. The ground truth distribution of values for this parameter is taken from the data, exported from Moldflow simulations.

The prediction is performed for a subset of previously randomly selected points. Then for each selected point the predicted value is copied over a selected neighboring area. Values for overlapping parts of areas are averaged. This allows us to create smooth distribution that reflects the continuous nature of the fill time value and reduces computational time.

\subsection{Fill time projection to 2d space}

For deflection prediction, we project a 3d point-cloud with the predicted fill time distribution to the 2d plane. The projection plane is selected so that it minimizes the squared Euclidean 2-norm for the original and projected points. The projected image is 384x768 pixels in size. As a feature map for the second channel, we also obtain a 2d map of a mask that represents the silhouette of the projection of the mesh, to help the model to distinguish the projection from the background. This approach allows us to take into account the geometrical features of the mesh alongside derivative features of the injector’s gates in the form of predicted fill time.

\subsection{Deflection prediction with the CNN model} \label{ch:deflection_prediction_algo}

For the prediction of the distribution of deflection values over the 2d image we use a convolution neural network model based on the U-Net model \cite{ronneberger2015unet}. The details of the architecture of the model are described in the appendix.

As a result of inference with our model, we obtain a 2d map of predicted distribution 12x24 in size. This 2d representation is then up-scaled, with interpolation, to represent the 2d image, 384x768  in size, of predicted deflection distribution over the initial 2d projection. Prediction of 2d deflection map in a lower resolution and then up-scaling it allows us to obtain a smooth result distribution that reflects the continuous nature of the deflection.

\subsection{Inflate 2d prediction data back to 3d mesh} \label{ch:deflection_inflate_2d}

The predicted 2d deflection map then is re-projected back onto 3d space to produce distribution over the initial point cloud. For each of 3d points, we pick up the corresponding value from the 2d plane using information about 3d-to-2d points correspondence saved on the projection step.

\section{Experiments and Results}
\subsection{Dataset}
We use the dataset obtained from our aforementioned partner, which consists, after extraction (described in section \ref{ch:data_preparation}), of 158 Moldflow injection molding simulations for car dashboards and bumpers. Each simulation consists of 3d mesh along with Moldflow simulation parameters and results. All simulations are performed using same type of polymer, without a cooling system. Statistics of mesh parameters from the dataset are shown in Table \ref{tab:meshes}.

\begin{table}[h]
    \centering
        \caption{Statistics of mesh characteristics}
    \begin{tabular}{cccc}
        \hline
        Parameter & Min & Median & Max \\
        \hline
        Vertices &  35311 & 77846 & 110126\\
        Edges & 212172 & 462420 & 662472 \\
        Faces & 70724 & 156140 & 220824 \\
        \hline
    \end{tabular}

    \label{tab:meshes}
\end{table}

Units of fill time values are measured in seconds (s). Deflection values are measured in millimeters (mm).

\subsection{Metrics}

To assess performance of our predictors we calculate RMSE and MAE metrics taking final values of predicted distribution over a 3d point cloud after all steps, described in chapter \ref{ch:filltime_prediction_algo} and \ref{ch:deflection_inflate_2d}. As a final result for each fold of cross-validation we present mean values of these metrics averaged over test samples and values of metrics calculated on all points from all of the test samples of the fold, merged into a single array.

\subsection{Cross-validation}

We trained and tested our predictors of fill time and deflection using cross-validation with five folds. Parameters of cross-validation are shown in Table \ref{tab:crossvalidaiton_parameters}.

\begin{table}[h]
    \centering
        \caption{Cross-validation parameters}
    \begin{tabular}{lc}
        \hline
        Parameter & Value \\
        \hline
        Total simulation samples & 158 \\
        Cross validation folds & 5 \\
        Train samples per fold  & 127 \\
        Test samples per fold & 31 \\
        \hline
    \end{tabular}

    \label{tab:crossvalidaiton_parameters}
\end{table}

Both of fill time and deflection predictors are trained and tested on the same split of the dataset for each of the cross-validation folds.

\subsection{Fill time prediction}

At the first stage, we predict the distribution of fill time with the Gradient Boosting method using XGBoost library. The hyperparameters used for XGBoost regressor training are shown in Table \ref{tab:xgboost_parameters}.

\begin{table}[h]
    \centering
        \caption{XGBoost regressor hyperparameters}
    \begin{tabular}{lc}
        \hline
        Parameter & Value \\
        \hline
        Learning rate & 0.08 \\

        Maximum depth & 8 \\

        Number of estimators & 200 \\

        Importance type & gain \\
        \hline
    \end{tabular}
    \label{tab:xgboost_parameters}
\end{table}

Table \ref{tab:filltime_results} shows the results of fill time prediction per each fold and mean values of metrics over all folds.

\begin{table}[!h]
    \centering
        \caption{Fill time prediction results}
    \begin{tabular}{lcccc}
        \hline
        \textbf{Fold}
            & \shortstack{\\ \textbf{RMSE} \\ samples}
            & \shortstack{\\ \textbf{MAE} \\ samples}
            & \shortstack{\\ \textbf{RMSE} \\ points}
            & \shortstack{\\ \textbf{MAE} \\ points} \\
        \hline
        \#1 & 0.7600 & 0.5831 & 0.8490 & 0.5234 \\
        \#2 & 0.5397 & 0.3840 & 0.6035 & 0.3781 \\
        \#3 & 0.6045 & 0.4571 & 0.6561 & 0.3897 \\
        \#4 & 0.5243 & 0.3762 & 0.5811 & 0.3597 \\
        \#5 & 0.6291 & 0.4698 & 0.7699 & 0.4543 \\
        \hline

    \multicolumn{5}{c}{\textbf{Summary}} \\
    \hline

        \textbf{Mean} & \textbf{0.6115} & \textbf{0.4540} & \textbf{0.6919} & \textbf{0.4210} \\
    \hline
    \end{tabular}
    \label{tab:filltime_results}
\end{table}

To train our CNN model we use samples with both predicted and ground truth fill time distributions with the same target 2d maps of deflection distribution. These maps are obtained from the ground truth deflection distribution with the same procedure of 2d projection, after which they are resized to the target dimensions of 12x24.


Also we augment the dataset by mirroring maps of each sample over the $x$ and $y$ axes.

\newpage
To test the deflection predictor we use the predicted fill time for the test sample and take average of four predicted 2d deflection maps for one original and three mirrored 2d fill time maps. After that we perform the reprojection of the results of 2d deflection prediction back to the 3d point cloud, as described above, and calculate the metrics for this resulting 3d distribution. Table \ref{tab:deflection_results} shows the results of deflection prediction.

\begin{table}[!h]
    \centering
        \caption{Deflection prediction results}
    \begin{tabular}{l c c c c }
        \hline
        \textbf{Fold}
            & \shortstack{\\ \textbf{RMSE} \\ samples}
            & \shortstack{\\ \textbf{MAE} \\ samples}
            & \shortstack{\\ \textbf{RMSE} \\ points}
            & \shortstack{\\ \textbf{MAE} \\ points} \\
        \hline
        \#1 & 2.3837 & 2.0582 & 2.6504 & 1.9954 \\
        \hline
        \#2 & 1.6406 & 1.2530 & 1.9034 & 1.2312 \\
        \hline
        \#3 & 1.1052 & 0.8111 & 1.2956 & 0.7779 \\
        \hline
        \#4 & 1.2364 & 0.8587 & 1.3846 & 0.8651 \\
        \hline
        \#5 & 3.1082 & 2.7538 & 3.2924 & 2.6459 \\
        \hline

    \multicolumn{5}{c}{\textbf{Summary}} \\
    \hline

        \textbf{Mean} & \textbf{1.8948} & \textbf{1.5469} & \textbf{2.1053} & \textbf{1.5031} \\
    \hline
    \end{tabular}
    \label{tab:deflection_results}
\end{table}

\subsection{Simulation time comparison}

We obtained the execution times for each phase of a simulation using our solution. The configuration of the test system is presented in Table \ref{tab:test_setup}.
\begin{table}[!h]
    \centering
        \caption{Test bench setup}
    \begin{tabular}{l c}
        \hline
            \textbf{CPU} & Intel Xeon E5-2630 v4 40 cores 2.20GHz \\
        \hline
            \textbf{RAM} & 125.8 GB \\

            \textbf{OS} & Ubuntu 16.04 LTS \\

            \textbf{GPU}  & GeForce RTX 2080 Ti \\
        \hline
    \end{tabular}
    \label{tab:test_setup}
\end{table}

We measured time for three main parts of the calculations: initial data pre-processing, fill time prediction and deflection prediction. The summary results are shown in Table \ref{tab:simulation_time}

\begin{table}[!h]
    \centering
        \caption{Summary of simulation time with our solution, in seconds}
    \begin{tabular}{lccccc}
        \hline
              \textbf{Stage}
            & \textbf{Min}
            & \textbf{Mean}
            & \textbf{Max}
            & \textbf{Std}
            & \textbf{Med}
            \\
        \hline
            \textbf{Pre-processing} & 13.611 & 95.961 & 224.120 & 66.317 & 76.022\\

            \textbf{Fill time} & 1.390 & 3.529 & 5.716 & 1.18 & 3.534 \\

            \textbf{Deflection} & 1.015 & 2.294 & 4.566 & 0.61 & 2.302 \\

            \textbf{Total} & \textbf{16.016} & \textbf{101.784} & \textbf{232.341} & \textbf{67.873} & \textbf{81.948} \\
        \hline
    \end{tabular}
    \label{tab:simulation_time}
\end{table}

For comparison with Moldflow performance we use execution time from log files from our dataset of simulations. Since these files provide only overall time for the Warp Analysis phase and for Fill Analysis combined with other types of analysis (for example, with both Pack Analysis and Residual Stress Analysis), if they are provided, we cannot perform direct comparison of times of isolated fill time and deflection prediction. Therefore, for the fill time phase only, we took only simulations that have execution time of Fill Analysis without conjunction with any other type of analysis. In our dataset Moldflow simulations were performed on different hardware, so to make the comparison more precise, we took only those simulations that were performed on a hardware configuration as close to ours as we could select (CPU of 40 and more cores).

There were seven such simulations in our dataset. The result is shown in Table \ref{tab:fill_analysis_only}.

\begin{table}[!h]
    \centering
        \caption{Execution time for Moldflow simulations with Fill analysis only, in seconds}
    \begin{tabular}{ccc}
        \hline
              \textbf{Min}
            & \textbf{Mean}
            & \textbf{Max}
            \\
        \hline
            291 & 869.143 & 1103 \\
        \hline
    \end{tabular}
    \label{tab:fill_analysis_only}
\end{table}

It is evident that our solution is faster in passing all the stages that are needed to compute fill time distribution, even if we compare the fastest Moldflow simulation, with an execution time of $291$ sec., with the slowest pre-processing and fill time prediction stages with our solution, which is $229.135$ sec.

For deflection prediction time we took simulations that contain a Warp analysis stage and were performed on the most similar hardware. There were 99 simulations selected. Then for each simulation we obtained the total time of execution of all stages of the simulation, including analysis stages preceding the Warp analysis (for example, the Fill analysis). Table \ref{tab:warp_analysis} shows the summary of the obtained time periods.

\begin{table}[h]
    \centering
        \caption{Total execution time summary for Warp analysis in selected Moldflow simulations, in seconds}
    \begin{tabular}{ccccccc}
        \hline
              \textbf{Min}
            & \textbf{Mean}
            & \textbf{Max}
            & \textbf{Std}
            & \textbf{25\%}
            & \textbf{Med}
            & \textbf{75\%}
            \\
        \hline
            127 & 1765.4 & 38116 & 4222.6 & 520 & 1161 & 1594.5 \\
        \hline
    \end{tabular}
    \label{tab:warp_analysis}
\end{table}

We compare these results with the total time of all stages to obtain deflection prediction with our solution. There are only nine Moldflow simulations, with a maximum execution time of $177$ sec. and a minimum of $127$ sec., that are faster than the maximum total time with our solution. These Moldflow simulations are with different technological parameters and the same geometry. And our solution gives execution time for these simulations in the interval from $29.075$ to $33.05$ sec., which is still multiple times faster.

On average our solution for deflection prediction appeared to be $1663.616$ sec. and $17.34$ ($14.17$ by median) times faster than Moldflow.

\section*{Conclusion}

In this work we proposed the baseline pipeline for data processing of injection molding simulation parameters to predict target distributions of deflection and fill time over a 3d mesh using a surrogate modelling approach. The pipeline includes the extraction of data from Moldflow simulation projects and the prediction of the target distributions.

We described how and where to get valuable data from files of injection molding simulation, extracted from Moldflow.

Then we proposed the algorithm for engineering of features, that will be used to train model for prediction of fill time distribution. Derived features include information about location and direction of injector’s gates that will mostly affect the time of plastic to reach the particular point of a 3d mesh, relative to each point of the point cloud of a mesh, alongside information about the opening time of gates.

As a model for fill time prediction we proposed using the Gradient boosting technique, and we showed its performance on our data using a five-fold cross-validation check and obtained the baseline values of MSE and RMSE metrics for the fill time distribution prediction.

Then we described the algorithm for engineering of features for predicting the deflection distribution that will include information about the geometry as well as parameters of gates in form of predicted fill time distribution.

For the deflection distribution we proposed a 2d convolutional neural network model and steps for features and target distibution projection to 2d dimensional space and back. We showed the performance of the proposed method using the same cross-validation split and obtained the baseline values of MSE and RMSE metrics for the deflection distribution prediction.

Finally we measured the execution time of our solution for the steps of targets prediction and compared it to the time of simulations with Moldflow software. The result shows that our solution significantly excels the Moldflow in execution time: for about 17 times faster comparing mean and for about 14 times comparing median for total time of all stages of the deflection prediction. The slowest fill time prediction with our solution is faster than the quickest with Moldflow. And our slowest deflection prediction is faster than the vast majority of Moldflow simulations with Warp analysis from our dataset, while for each individual simulation with our solution is still multiple times faster.

To demonstrate the potential of the described solution, we developed a web application prototype that allows a user to upload molding data of a 3d mesh, set up injector's gates parameters, perform simulation using our data preprocessing algorithms and trained ML models, and obtain vivid visualization of its result. This prototype was presented and approved by the management board of the Fiat Chrysler Automobiles, the Italian automotive manufacturer.

In the further work, we will aim to improve the quality of the model by the engineering of more features from the available data and testing the applicability of geometrical models for our task. We will also train our models on more data as it becomes available.

One of the nearest ways of improvement of our current model, we see using a multiple view approach and including information of distance from 3d point to its projection as an additional geometrical feature. Another direction of model enhancement is to use ANN models that work directly with Point Cloud and Mesh or Graph representations, such as DGCNN \cite{wang_dynamic_2019}.

Finally, we see one of the possible but promising applications of such surrogate modelling approach in using trained models as a fast objective function in the task of optimization of technological parameters (i.e. optimal gates placement), which could significantly help engineers in this task, or even automate it.

\section*{Acknowledgements}

\setcounter{footnote}{1}
\footnotetext{Skolkovo Institute of Science and Technology, Moscow, Russia}

We thank Vadim Leshchev$^{1}$ for his great contributions to the preparation of results presentation and technical implementation of ideas of this work, programming tools, and MVP application. We thank Ilnur Nuriakhmetov$^{1}$ for his great help in technical support of this project. We also thank Anna Nikolaeva$^{1}$ for her great help in the scientific research and the scientific report preparation. We thank Roman Misiutin for his great help in scientific research into graph models and this paper preparation.

\bibliographystyle{elsarticle-num}
\bibliography{refs.bib}

\newpage
\appendix
\section{Details of convolutional neural network architecture}

For the prediction of 2d deflection distribution we use slightly modified encoder half of the U-Net architecture. As the result it produces low-resolution 2d map of size 12x24 values. Table \ref{tab:unet_architecture} shows the details of the model structure. 
\begin{table*}[!h]
\begingroup
\setlength{\tabcolsep}{10pt} 
\renewcommand{\arraystretch}{1.5} 
    
    \caption{The structure of U-Net-based model for prediction of deflection distribution}
    \begin{tabular}{|c|c|c|c|}
        \hline
        Layer (type) & Output Shape & Params & Connected to \\
        \hline
        
        inputs\_img (InputLayer) & (384, 768, 2) & 0 &  \\
        \hline
        lambda (Lambda) & (384, 768, 2) & 0 & inputs\_img \\
        \hline
        conv2d (Conv2D) & (384, 768, 8) & 408 & lambda \\
        \hline
        conv2d\_1 (Conv2D) & (192, 384, 8) & 1608 & conv2d \\
        \hline
        conv2d\_2 (Conv2D) & (192, 384, 16) & 3216 & conv2d\_1 \\
        \hline
        conv2d\_3 (Conv2D) & (96, 192, 16) & 6416 & conv2d\_2 \\
        \hline
        conv2d\_4 (Conv2D) & (96, 192, 32) & 4640 & conv2d\_3 \\
        \hline
        conv2d\_5 (Conv2D) & (48, 96, 32) & 9248 & conv2d\_4 \\
        \hline
        conv2d\_6 (Conv2D) & (48, 96, 64) & 18496 & conv2d\_5 \\
        \hline
        conv2d\_7 (Conv2D) & (24, 48, 64) & 36928 & conv2d\_6 \\
        \hline
        conv2d\_8 (Conv2D) & (24, 48, 64) & 36928 & conv2d\_7 \\
        \hline
        conv2d\_9 (Conv2D) & (12, 24, 64) & 36928 & conv2d\_8 \\
        \hline
        conv2d\_10 (Conv2D) & (12, 24, 64) & 36928 & conv2d\_9 \\
        \hline
        conv2d\_11 (Conv2D) & (6, 12, 64) & 36928 & conv2d\_10 \\
        \hline
        conv2d\_transpose & & & \\
        (Conv2DTranspose) & (12, 24, 32) & 18464 & conv2d\_11 \\
        \hline
        concatenate (Concatenate) & (12, 24, 96) & 0 & conv2d\_transpose, conv2d\_9 \\
        \hline
        conv2d\_12 (Conv2D) & (12, 24, 32) & 27680 & concatenate \\
        \hline
        conv2d\_13 (Conv2D) & (12, 24, 32) & 9248 & conv2d\_12 \\
        \hline
        conv2d\_14 (Conv2D) & (12, 24, 1) & 289 & conv2d\_13 \\
        \hline
        conv2d\_15 (Conv2D) & (12, 24, 1) & 10 & conv2d\_14 \\
        \hline
        lambda\_1 (Lambda) & (12, 24, 1) & 0 & conv2d\_15 \\
        \hline
    \end{tabular}
    \label{tab:unet_architecture}
    \endgroup
\end{table*}

\end{document}